\pdfoutput=1

\documentclass[11pt]{article}

\usepackage[final]{acl}

\usepackage{times}
\usepackage{latexsym}

\usepackage[T1]{fontenc}

\usepackage[utf8]{inputenc}

\usepackage{microtype}

\usepackage{inconsolata}

\usepackage{graphicx}

\usepackage{{booktabs}}
\usepackage{tabularray}
\usepackage{multirow}
\usepackage{floatrow}

\usepackage[titlenotnumbered,ruled,vlined,noend,linesnumbered]{algorithm2e}
\usepackage{caption}

\title{An Incremental Clustering Baseline for Event Detection on Twitter}

\author{
  \textbf{Marjolaine Ray\textsuperscript{1}},
  \textbf{Qi Wang\textsuperscript{1}},
  \textbf{Frédérique Mélanie-Becquet\textsuperscript{1}},
  \textbf{Thierry Poibeau\textsuperscript{1}},
\\
  \textbf{Béatrice Mazoyer\textsuperscript{2}}
\\
\\
  \textsuperscript{1}Lattice (CNRS \& \'Ecole normale supérieure-PSL \& U. Sorbonne nouvelle), Paris, France \\
  \textsuperscript{2}médialab, Sciences Po, Paris, France
\\
  \small{
    \textbf{Correspondence:} \href{mailto:beatrice.mazoyer@sciencespo.fr}{beatrice.mazoyer@sciencespo.fr}
  }
}

\begin{document}
\maketitle
\begin{abstract}

Event detection in text streams is a crucial task for the analysis of online media and social networks. One of the current challenges in this field is establishing a performance standard while maintaining an acceptable level of computational complexity. In our study, we use an incremental clustering algorithm combined with recent advancements in sentence embeddings. Our objective is to compare our findings with previous studies, specifically those by \citet{cao_hierarchical_2024} and \citet{mazoyer_french_2020}. Our results demonstrate significant improvements and could serve as a relevant baseline for future research in this area.
\end{abstract}

\section{Introduction}

With the development of social media, the ability to recognize events in streams of short texts—particularly tweets—has become increasingly important. This process, called event recognition, involves identifying significant occurrences within large volumes of data, posing various challenges. A key component of this task is defining a clear and operational concept of what qualifies as an event. In this paper, we will use a working definition of event, as proposed by \citet{mcminn_building_2013}. The authors propose in fact  a double definition: ``\textbf{Definition 1}: An event is a \textbf{significant} thing that happens at some specific time and place''. This needs to be completed by the definition of what `significant' means, so they add: ``\textbf{Definition 2}: Something is significant if it may be discussed in the media. For example you may read a news article or watch a news report about it''. Because this definition has been used to build other corpora, it can be considered functional. As a result, corpora created with this definition should be comparable, with different annotators likely producing similar outcomes. 


One of the main challenges in event recognition is then being able to cluster different texts that refer to the same event. This difficulty arises from the wide range of expressions used to describe similar events. Different sources and users may refer to the same event using different expressions, making it essential for recognition systems to account for synonymy, paraphrasing, and other linguistic variations. Moreover, the temporal dimension is also a critical parameter in event recognition. The timing of events and the sequence in which they are reported can significantly impact the interpretation and relevance of the information extracted. Another challenge in event detection on social networks is the sheer volume of messages posted on these platforms: an effective algorithm must be capable of processing millions of tweets within a reasonable time frame. Many studies propose computationally intensive methodologies that are impractical for many real-world applications. Therefore, research in this field needs to establish baselines on publicly accessible datasets that are both performant and time-efficient.

The primary objective of this paper is thus to establish a performance standard for event detection while maintaining an acceptable level of computational complexity. Our approach involves the use of an incremental clustering algorithm enhanced by recent advancements in sentence embeddings.
Specifically, we build upon the incremental clustering algorithm introduced by \citet{mazoyer_french_2020} in their dataset publication. While effective at the time, their approach relied on lexical descriptions that may now be outdated due to the development of new word embedding techniques, particularly those stemming from recent large language models based on the transformer architecture \citep{vaswani2017attention}. In our study, we utilize Sentence-BERT \citep{reimers2019sentence}, a model that is especially noteworthy for its ability to encode entire sentences from individual word encodings.

The structure of the paper is as follows: First, we will review recent work in the domain. Next, we will detail our method and experiments. Finally, we will present and discuss our results, concluding with a broader discussion.
We conduct experiments on two large public Twitter datasets to demonstrate the state of the art performance, efficiency, and robustness of this method (note that our code is publicly accessible\footnote{\href{https://github.com/medialab/twitter-incremental-clustering/blob/main/README.md}{https://github.com/medialab/twitter-incremental-clustering/}}). We then aim to compare our results with previous studies in the field, specifically those by \citet{cao_hierarchical_2024} and \citet{mazoyer_french_2020}. By leveraging these advanced sentence embeddings, we demonstrate that our implementation surpasses more recent and complex approaches in both time-efficiency and the quality of detected events. These short-text representations provide a sophisticated understanding of language and context, allowing for more accurate and nuanced event recognition.

\section{Related Work}

\citet{hasan_survey_2018} conducted a comprehensive review of event detection techniques on Twitter. Like these authors, we identify three main categories of methods: ‘term-interestingness-based’ approaches, topic modeling, and incremental clustering. However, we expand upon their typology by adding a fourth category: graph-based approaches.

\paragraph{``Term-Interestingness-Based'' Approaches.}
These methods involve monitoring terms that are probably associated with an event, often identified by a sudden increase in the frequency of certain terms. Typically, they return the top trending events on Twitter. These approaches generally do not allow the detection of low-bursty events.

\paragraph{Topic Modelling.}
Topic models are widely used techniques derived from Latent Dirichlet Allocation (LDA) \citep{blei_latent_2003} to uncover the thematic structure within a collection of textual documents. Several works have been interested in adapting this method to make topics evolve over time, and to adapt to the short format of tweets by restricting the number of topics associated with a document. \citet{likhitha_detailed_2019} propose a survey of topic modeling methods adapted to short texts.

\paragraph{Incremental Clustering.}
This family of methods derives from the Topic Detection and Tracking (TDT) initiative \citep{allan_topic_1998}, aimed at identifying and following events in a stream of broadcast news stories. The task of detecting new events (First Story Detection) involves representing documents as vectors in a semantic space. Each new document is compared to existing ones (or to a set of past documents within a time-window) and if its similarity to the closest document (or centroid) falls below a defined threshold, it is identified as a new story. This methodology was then adapted to event detection on Twitter \citep{petrovic2010streaming, mcminn2015real} with tf-idf \citep{sparck_statistical_1972} as a vector representation of tweets. More recent works \citep{mazoyer_french_2020, qiu_single_2021, pradhan_edtbert_2024} use BERT \citep{devlin_bert_2019} or Sentence Transformers \citep{reimers2019sentence} to produce a vector representation of tweets.

\paragraph{Graph-Based Approaches.}
These methods \citep{peng2022reinforced, ren2022known, cao_hierarchical_2024} leverage the semantic structure of social media, using anchors such as hashtags, user mentions, hyperlinks and named entities. They construct message graphs that include all candidate messages, linking those that share common attributes. The event detection task is then framed as a graph-partitioning problem.

\section{Methodology}

When working with social media data, one needs to consider both the textual similarity of the documents and their temporal proximity to avoid grouping together tweets posted at significantly different times. Since the number of events is not known in advance, the chosen algorithm does not require the number of events given a priori. Following the method by \citet{mazoyer_french_2020}, we use an incremental clustering algorithm derived from the Topic Detection and Tracking \citep{allan_topic_1998} initiative.
\paragraph{Algorithm.}
\begin{algorithm*}
\SetKwData{Left}{left}\SetKwData{This}{this}\SetKwData{Up}{up}
\SetKwFunction{Union}{Union}\SetKwFunction{FindCompress}{FindCompress}
\SetKwInput{Input}{input}\SetKwInOut{Output}{output}
\SetKwBlock{DoParallel}{do in parallel}{end}
\Input{threshold $t$, window size $w$, batch size $b$, corpus $C=\{d_0 \ldots d_{n-1}\}$ of $n$ documents in chronological order}
\Output{a list $T$ of cluster ids for each document}
\BlankLine
$T \leftarrow \left[ \right] ; i \leftarrow 0  ; j \leftarrow 0 $\;
\While{$i < n - b $}{
$batch = \{d_{i}, \ldots d_{i + b -1}\}$\;
\DoParallel{
\For{document $d$ in $batch$}{
\eIf{$T$ is empty}{
$cluster\_id(d) \leftarrow j$\;
$j \leftarrow j+1$\;
}{
$d_{nearest} \leftarrow $ nearest neighbor of $d$ in $T$\;
\eIf{$\delta(d, d_{nearest}) < t$}{
$cluster\_id(d) \leftarrow cluster\_id(d_{nearest})$\;
}{
$cluster\_id(d) \leftarrow j$\;
$j \leftarrow j+1$\;
}
}
\If{$|T| \geq w$\;}{remove first document from $T$}

add $d$ to $T$\;
}
}
$i \leftarrow i + b$\;
}
\caption{``mini-batch" FSD}\label{FSD_mini}
\end{algorithm*}
This mini-batch First Story Detection (FSD) algorithm works as follows: documents are vectorized (we develop embedding methods in the subsequent section), sorted chronologically, and processed in batches of $b$ documents. Each new batch is compared to a window of $w$ previous documents in terms of cosine distance. For each batch document, if the distance to its nearest neighbor is smaller than a threshold $t$, it joins the same cluster as its nearest neighbor. Otherwise, the document joins a new cluster. The procedure is detailed in Algorithm~\ref{FSD_mini}, where $\delta$ denotes the cosine distance.

\paragraph{Short-Text Embeddings.}
In the work published by \citet{mazoyer_french_2020}, the best performing embedding method is a tf-idf score where the $df$ (document-frequency) is computed over the entire tweet dataset (millions of tweets). Over the past five years, numerous models have emerged, particularly large language models (LLMs), which are especially suited for this task as they encode both linguistic and world knowledge, making them highly effective in capturing the nuances and complexities of event detection. We use Sentence Transformers, also known as SBERT \citep{reimers2019sentence}, a BERT/RoBERTa \citep{devlin_bert_2019, liu2019roberta} fine-tuning architecture using Siamese networks. This model ensures that the resulting sentence embeddings are both semantically meaningful and comparable, using cosine distance.

\begin{figure}[t]
  \includegraphics[width=\columnwidth]{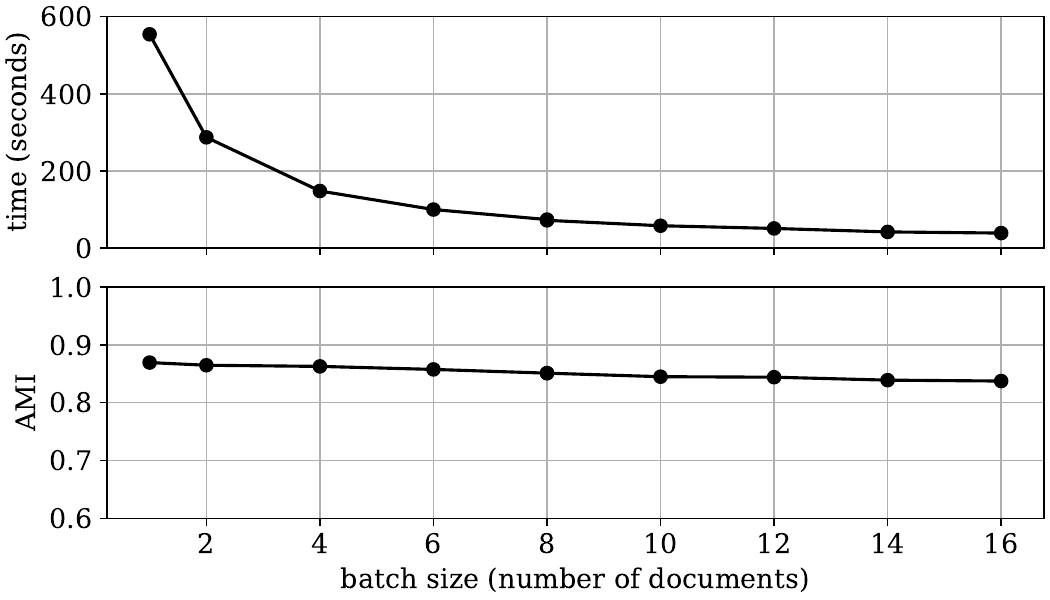}
  \caption{Evolution of execution time and adjusted mutual information (AMI) of the "mini-batch" FSD algorithm depending on batch size $b$ on the entire Event2012 corpus (68,841 documents).}
  \label{fig:timeplot}
\end{figure}

\paragraph{Time Complexity.}
The time complexity of the FSD algorithm is $O(nw)$ (with $n$ the number of documents in the collection and $w$ the number of documents in the time window), since each document in the corpus is compared only with the last $w$ documents in chronological order. In practice, when using the "mini-batch" FSD, computation time is inversely proportional to batch size, as illustrated in Figure~\ref{fig:timeplot}.

\section{Experiments}   

\paragraph{Baselines.}
We compare our results (\textbf{FSD-SBERT}) with HISEvent\footnote{\href{https://github.com/SELGroup/HISEvent}{https://github.com/SELGroup/HISEvent}} (\textbf{HE}), the most recent paper on event detection: \citeposs{cao_hierarchical_2024} work on the partition of a graphical neural network representation of tweets using structural entropy minimization. We also evaluate the performance improvement achieved by using Sentence Transformers in comparison to the tf-idf vectors used in (\textbf{TW})\footnote{\href{https://github.com/ina-foss/twembeddings}{https://github.com/ina-foss/twembeddings}} by \citet{mazoyer_french_2020}.

\paragraph{Datasets.}
We conducted experiments using two extensive, publicly accessible tweets datasets: Event2012 \citep{mcminn_building_2013} and Event2018 \citep{mazoyer_french_2020}. The Event2012 dataset contains 150,000 English tweet IDs related to 506 distinct events over a four-week period. In contrast, Event2018 comprises 96,000 French tweet IDs corresponding to 257 unique events, all posted within a span of 23 days. For a fair comparison with baseline methods, we limit our analysis to the subset of the dataset used by  \citet{cao_hierarchical_2024}. Indeed, these authors downloaded the tweets recently after many were deleted. Their dataset, therefore, contains 68,841 tweets related to 503 events for Event2012 and 64,516 tweets related to 257 events for Event2018. We do not use the distinction adopted by \citet{cao_hierarchical_2024} between open-set (day-by-day detection) and closed-set (detection across the entire corpus), as we argue that events should be allowed to span multiple consecutive days. Therefore, we only evaluate our method on the complete corpus.

\paragraph{Short-Text Embeddings.} We use Sentence Transformers (SBERT) models pre-trained on English and French corpora to compute vectors from tweets. Specifically, we use \textbf{all-mpnet-base-v2}\footnote{ \href{https://huggingface.co/sentence-transformers/all-mpnet-base-v2}{https://huggingface.co/sentence-transformers/all-mpnet-base-v2}} for the English dataset and \textbf{Sentence-CamemBERT-Large}\footnote{ \href{https://huggingface.co/dangvantuan/sentence-camembert-large}{https://huggingface.co/dangvantuan/sentence-camembert-large}} \citep{martin2020camembert} for the French dataset.

\paragraph{Parameters.} The mini-batch FSD algorithm takes three input parameters: the cosine distance threshold ($t$), the time-window size ($w$) and the batch-size ($b$). Consistently with \citet{mazoyer_french_2020}, we set $w$ to the average number of documents per day in each dataset, and the batch size to 8 documents. The threshold $t$ depends on the type of text-embedding. It was optimized using grid-search and set to $0.5$ for English and $0.55$ for French.

\paragraph{Evaluation Metrics.}
\begin{table}
\centering
\begin{tabular}{llccc}
\toprule
 &  & \textbf{FSD-SBERT} & \textbf{HE} & \textbf{TW} \\
dataset &  &  &  &  \\
\midrule
\multirow[c]{2}{*}{\textbf{2012}} & \textbf{ARI} & \bfseries 0.63 & 0.50 & 0.39 \\
 & \textbf{AMI} & \bfseries 0.86 & 0.81 & 0.82 \\
 \midrule
\multirow[c]{2}{*}{\textbf{2018}} & \textbf{ARI} & \bfseries 0.55 & 0.44 & 0.25 \\
 & \textbf{AMI} & \bfseries 0.81 & 0.66 & 0.72 \\
\bottomrule
\end{tabular}

\caption{ARI and AMI scores on two datasets: Event2012 (in English) and Event2018 (in French).}
\label{tab:ami_ari_metrics}
\end{table}

We use the scikit-learn \citep{pedregosa_scikit-learn_2011} implementation of adjusted mutual information (AMI) \citep{vinh2009information} and adjusted rand index (ARI) \citep{rand1971objective}, which are widely employed in event detection evaluation \cite{cao_hierarchical_2024}. 

\section{Results and Discussion}

\paragraph{Performance.} Table~\ref{tab:ami_ari_metrics} compares the performance of our method (\textbf{FSD-SBERT}) with \textbf{HE} and \textbf{TW}. We observe that  \citeposs{mazoyer_french_2020} mini-batch FSD algorithm combined with Sentence Transformers pre-trained on large text corpora consistently outperforms the baselines on both datasets. The comparison between HISEvent (HE) and twembeddings (TW) seems to indicate that the mini-batch First Story Detection algorithm, even used with a simple tf-idf representation of tweets, is still a strong baseline, since its performance is comparable (and even superior on the French dataset) to HISEvent when using AMI as the indicator, though it is inferior when evaluated with ARI.

\begin{figure}[t]
  \includegraphics[width=\columnwidth]{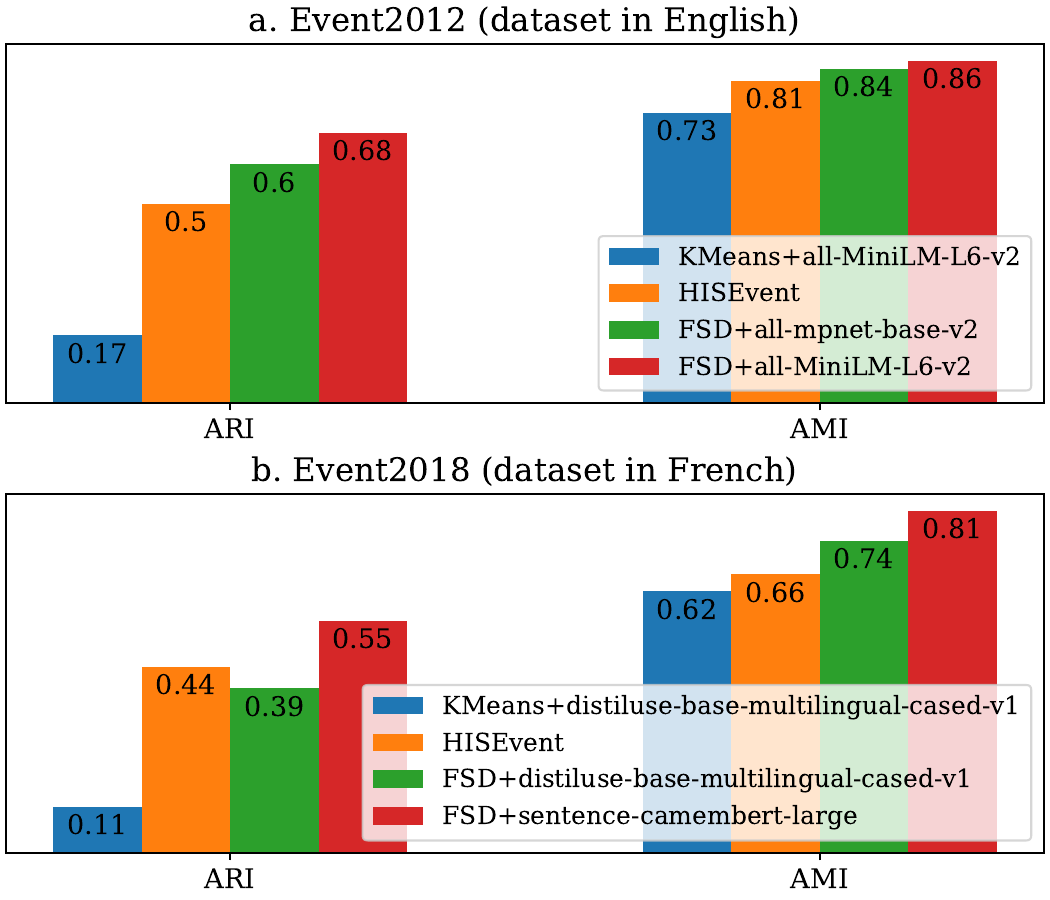}
  \caption{ARI and AMI scores with different SBERT models and different clustering algorithms. All FSD tests ran with $b=8$ and $t=0.55$.}
  \label{fig:sbertplot}
\end{figure}

It is important to note that \citet{cao_hierarchical_2024} also use Sentence Transformers as a baseline in their article, with a different clustering algorithm (K-means). Their results are represented as the first column in Figure~\ref{fig:sbertplot}, with the exact SBERT models they have used ("all-MiniLM-L6-v2"\footnote{\href{https://huggingface.co/sentence-transformers/all-MiniLM-L6-v2}{https://huggingface.co/sentence-transformers/all-MiniLM-L6-v2}} for English and "distiluse-base-multilingual-cased-v1"\footnote{\href{https://huggingface.co/sentence-transformers/distiluse-base-multilingual-cased-v1}{https://huggingface.co/sentence-transformers/distiluse-base-multilingual-cased-v1}} for French). Our experiments show that the type of SBERT model has an effect on performance: as shown on Figure~\ref{fig:sbertplot} b, the "multilingual" model is less efficient for French than the language-specific "CamemBERT" model.  Nevertheless, regardless of the model used, the FSD algorithm (see the last two columns) is much more efficient than the K-means for both datasets. This gap is explained by the fact that the FSD algorithm is able to take into account the temporality of tweets (by applying a sliding time window when searching for nearest neighbors) unlike the K-means. Moreover, FSD seems to be robust to changes in SBERT models without the need to adapt the parameters: on Figure~\ref{fig:sbertplot}, when using FSD, the same threshold $t=0.55$ is used for all SBERT models. This common threshold explains the small difference between the values in  Table~\ref{tab:ami_ari_metrics} and Figure~\ref{fig:sbertplot} for the all-mpnet-base-v2 model, since the threshold is set to $0.5$ in Table~\ref{tab:ami_ari_metrics} and to $0.55$ in Figure~\ref{fig:sbertplot}.

\paragraph{Time efficiency.}
Increasing the batch size is a way to increase the computation speed with minimal loss in clustering performance: as shown in Figure~\ref{fig:timeplot}, doubling the batch size only decreases the performance (measured by AMI) by 0.5\%. This is why our experiments were all run with a batch size ($b$) set to 8 documents. With these parameters, our algorithm processes the Event2012 corpus, consisting of 68,841 documents (with a window size $w$ of 2,368 documents), in 72 seconds. In contrast, HISEvent requires 1 hour and 45 minutes to process a block of 8,722 documents, and over 5 days to handle the entire corpus.

The experiments shown on Figure~\ref{fig:timeplot} were run on a notebook PC with 32GB of RAM and and 8 2.4GHz CPUs. Note that these tests do not take into account the encoding of the tweets using Sentence Transformers, since we computed the embeddings only once on a GPU server and then stored them to be re-used for further experiments on a notebook computer without GPU. It took 65 seconds using a NVIDIA RTX A4500 GPU to encode the Event2012 corpus, and 240 seconds to encode the Event2018 corpus.

\paragraph{Resources.} We observed that executing HISEvent on the entire Event2012 dataset required substantial memory resources, exceeding 62 GB of RAM. In contrast, FSD operates with significantly lower memory requirements (less than 32GB of RAM).

\paragraph{Limitations.} 
Twitter has been an invaluable resource for research on social media and real-time data streams. However, this is no longer possible due to the platform’s API restrictions. Nevertheless, we believe this study remains relevant, as other data streams and social networks continue to produce valuable data, and event recognition continues to be a crucial task.

Another limitation related to the mini-batch FSD algorithm is the need to pre-determine the hyper-parameter $t$. However, the consistency of the results with the same $t$ value across several SBERT models (see Figure~\ref{fig:sbertplot}) suggests that this threshold ($t=0.55$) could be appliced to other Sentence Transformers models pre-trained on corpora in different languages.

Finally, a potential improvement for this method would be to better account for the nested nature of events in public discourse: for instance, a major political event might consist of numerous smaller sub-events, such as speeches, protests, and negotiations (for example the Yellow Vest protest in France lasted several months, with protests every week, discussions with the government, thousands of declarations, actors and reactions \cite{wagner2022}. Each of these sub-events can be reported separately (or not) in different messages. This layered structure would ideally necessitate more sophisticated models capable of capturing and integrating these various components to provide a coherent and comprehensive understanding of the overall event.

\section{Conclusion}

In this study, we aimed to investigate the performance of incremental clustering combined with Sentence Transformers models for  automatically detecting events in a stream of tweets. Our results demonstrated that applying the mini-batch FSD algorithm to SBERT representations significantly improves event detection performance on Twitter. We suggest that future research in this area should adopt this straightforward approach as a baseline for deploying more complex algorithms.

\section*{Acknowledgements}
This work was funded in part by the Agence Nationale de la Recherche, as part of the ANR Medialex Project (AAPG 2021). Thierry Poibeau is also funded by the “Investissements d’avenir" program, reference ANR-19-P3IA-0001 (PRAIRIE 3IA Institute).

\bibliography{custom}

\end{document}